\documentclass[10pt,twocolumn,letterpaper]{article}

\usepackage{cvpr}              

\usepackage[dvipsnames]{xcolor}

\definecolor{cvprblue}{rgb}{0.21,0.49,0.74}
\usepackage[pagebackref,breaklinks,colorlinks,citecolor=cvprblue]{hyperref}
\usepackage{bm}
\usepackage{siunitx}
\usepackage{wrapfig}
\usepackage{stmaryrd}
\usepackage{amsfonts}
\usepackage{multirow}
\usepackage[accsupp]{axessibility} 
\newtheorem{definition}{Definition}

\def\mygreen#1{\textcolor{green!65!black}{#1}}
\def\myred#1{\textcolor{red!65!black}{#1}}

\def\paperID{3887} 
\def\confName{CVPR}
\def\confYear{2024}

\title{Align and Aggregate: Compositional Reasoning with Video Alignment and Answer Aggregation for Video Question-Answering}

\author{Zhaohe Liao$^1$\thanks{These two authors contributed equally to this work.}\\
\and
Jiangtong Li$^2$\footnotemark[1]\\
\and
Li Niu$^1$\thanks{The corresponding authors.}\\
\and
Liqing Zhang$^1$\footnotemark[2]\\
\and
$^1$ Shanghai Jiao Tong University\\
    {\tt\small \{zhaoheliao, ustcnewly, zhang-lq\}@sjtu.edu.cn}\and
$^2$ Tongji University\\{\tt\small jiangtongli@tongji.edu.cn}
}

\begin{document}
\maketitle
\begin{abstract}
    Despite the recent progress made in Video Question-Answering (VideoQA), these methods typically function as black-boxes, making it difficult to understand their reasoning processes and perform consistent compositional reasoning.
    To address these challenges, we propose a \textit{model-agnostic} Video Alignment and Answer Aggregation (VA$^{3}$) framework, which is capable of enhancing both compositional consistency and accuracy of existing VidQA methods by integrating video aligner and answer aggregator modules.
    The video aligner hierarchically selects the relevant video clips based on the question, while the answer aggregator deduces the answer to the question based on its sub-questions, with compositional consistency ensured by the information flow along question decomposition graph and the contrastive learning strategy.
    We evaluate our framework on three settings of the AGQA-Decomp dataset with three baseline methods, and propose new metrics to measure the compositional consistency of VidQA methods more comprehensively.
    Moreover, we propose a large language model (LLM) based automatic question decomposition pipeline to apply our framework to any VidQA dataset.
    We extend MSVD and NExT-QA datasets with it to evaluate our VA$^3$ framework on broader scenarios.
    Extensive experiments show that our framework improves both compositional consistency and accuracy of existing methods, leading to more interpretable real-world VidQA models.
\end{abstract}    

    \section{Introduction}\label{sec:introduction}
    Video Question-Answering (VidQA) has emerged as a popular research topic in recent years, with potential applications in interactive artificial intelligence and recognition science.

    With the development of representing video, question and their alignment, numerous works~\cite{GaoGCN18, JiangCLZG20, HuangCZDTG20, XiaoYL0JC22, 0004WXJC22, 00040XC22, CherianHMR22} have achieved considerable success in both open-ended VidQA~\cite{XuMYR16, JangSYKK17} and multi-choice VidQA~\cite{JangSYKK17, XiaoSYC21, Li0022}.

\begin{figure}
        \centering
        \includegraphics[width=0.9\linewidth]{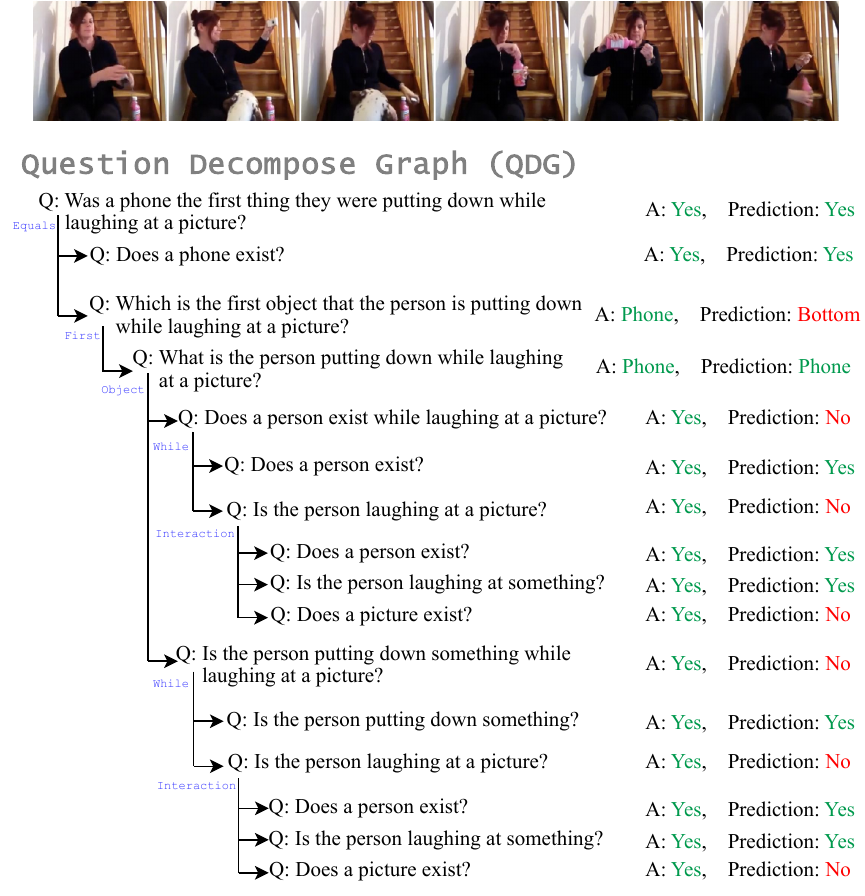}
        \caption{
            The Question Decomposition Graph (QDG) of a question from AGQA-Decomp~\cite{gandhi2022measuring}.
             The predicted answer to each question is from HQGA~\cite{XiaoYL0JC22}.
             Green (\emph{resp.}, Red) represents the predicted answer is right (\emph{resp.}, wrong)
        }
        \label{fig:example}
    \end{figure}

    However, existing VidQA methods often function as black-box models, making it difficult to understand the reasoning process behind their predictions and leading to inconsistent compositional reasoning.
    For example, in \Cref{fig:example}, HQGA~\cite{XiaoYL0JC22} can answer the question ``\textit{Is a phone the first object that the person is touching after taking a picture?}'' as ``\textit{Yes}''. However, HQGA can neither clearly identify the video clips that contain ``\textit{touch a phone}'' or ``\textit{take a picture}'' nor predict all the sub-questions correctly.
    Therefore, the lack of reasoning transparency can lead to poor compositional consistency, which reveals limited compositional reasoning ability, and further limits the accuracy of VidQA models, particularly on questions that involve temporal relations and multiple visual clues~\cite{gandhi2022measuring}.

    To tackle this issue, we introduce the Video Alignment and Answer Aggregation (VA$^{3}$) framework, which addresses these challenges by improving their compositional consistency and accuracy.
    This framework is \textit{model-agnostic} and can be applied to various VidQA methods, such as memory-based~\cite{GaoGCN18, FanZZW0H19}, graph-based~\cite{JiangH20, ParkLS21, Liu0WL21, GuZJHW21, WangBX22, CherianHMR22,SeoKPZ20}, and hierarchy-based~\cite{LeLV020, GuoZJ0L20, PengYBW21, DangLL021, PengWG0Z22, XiaoYL0JC22, abs220705342} methods.
    In detail, our VA$^{3}$ framework includes two additional modules, the video aligner and answer aggregator.
    The video aligner hierarchically aligns the question with the video clips from the object-level, appearance-level to motion-level.
    The answer aggregator takes the questions from the same Question Decomposition Graph (QDG) as input and deduces their answers based on their video-question joint representation.
    To the enhance compositional consistency, we further explore a contrastive learning strategy on the edge type of QDG.
    Overall, the VA$^{3}$ framework improves both compositional consistency and accuracy of existing VidQA methods, provides a more transparent compositional reasoning process, and further leads to more interpretable VidQA models in real-world applications.

    As for the evaluation metrics, AGQA-Decomp~\cite{gandhi2022measuring} proposes the compositional accuracy (CA), right for the wrong reasons (RWR), and delta (CA $-$ RWR) system to evaluate the compositional consistency of VidQA methods.
    However, these metrics only focus on reasoning failure based on the sub-questions correctness without considering the main question correctness, leading to asymmetric and unstable problems.
    To address this, we extend it to provide a symmetric and stable measurement for compositional consistency.
    In detail, our metrics include consistency precision~(cP), consistency recall~(cR), and consistency $\text{F}_1$~(c-$\text{F}_1$) along with their negative versions.
    These metrics can evaluate the compositional consistency of VidQA methods from a balanced viewpoint.
    More details are in~\Cref{sec:metrics}.

    We conduct the experiments on the AGQA-Decomp dataset~\cite{gandhi2022measuring} to verify the effectiveness of our framework.
    This dataset decomposes the questions into the sub-questions and the directed acyclic graphs, \ie the QDGs, making it applicable to evaluate the compositional consistency for VidQA methods.
    To validate the effectiveness and compositional consistency of our VA$^{3}$ framework, we conduct comprehensive experiments with three baseline methods: HME~\cite{FanZZW0H19}, HGA~\cite{JiangH20}, and HQGA~\cite{XiaoYL0JC22}, {which are the representations of the memory-based, graph-based and hierarchy-based methods, } in three different settings: balanced, novel compositions, and more compositional steps.
    Moreover, we propose an automatic question decomposition pipeline for VidQA datasets with the help of large language models (LLMs) to generalize our framework to datasets that do not have QDGs (\eg, MSVD~\cite{XuZX0Z0Z17} and NExT-QA~\cite{XiaoSYC21}) to verify the applicability of our framework.
    Additionally, we visualize the aligned video clips and the variation of predicted answers while equipping video aligner and answer aggregator successively to the backbone model on QDG to verify the interpretability of our framework.
    Our contribution can be summarized as:
    \begin{itemize}
        \item Dataset: We propose an automated question decomposition pipeline for any VidQA dataset to generate the QDGs and the sub-questions with the help of LLMs and further extend MSVD and NExT-QA dataset with it.
        \item Framework: We propose a \textit{model-agnostic} VA$^{3}$ framework, which provides a more transparent compositional reasoning process and increases both the interpretability and the accuracy of existing VidQA models.
        \item Metric: We extend the compositional consistency metrics as consistency precision (cP),
        consistency recall (cR) and consistency $\text{F}_1$ (c-$\text{F}_1$) along with their negative versions for a more balanced and comprehensive evaluation.
        \item Experiments:
        Comprehensive experiments with three baselines on five benchmark settings of three datasets reveal that our framework significantly boosts these baselines in compositional consistency and accuracy.
    \end{itemize}

    \section{Related Work}
    \begin{figure*}[ht]
        \centering
        \includegraphics[width=0.95\linewidth]{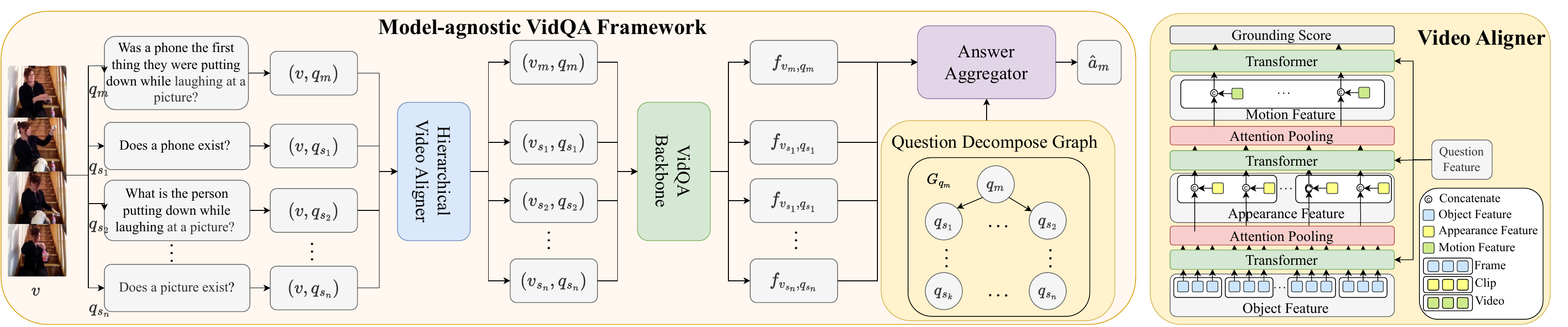}
        \caption{
            Our \textit{model-agnostic} Video Alignment and Answer Aggregation (VA$^3$) framework.
            $(v, q)$ is a video-question pair, where $q_m$ denotes main-question and ${q_{s_1}, \cdots, q_{s_n}}$ denote the $n$ sub-questions derived from $q_m$.
            $v_m$ and $\{v_{s_1}, \cdots, v_{s_n}\}$ denote the aligned videos according to corresponding questions.
            $\bm{f}_{v_m, q_m}$ and $\{\bm{f}_{v_{s_1}, q_{s_1}}, \cdots, \bm{f}_{v_{s_n}, q_{s_n}}\}$ denote the video-question joint features.
            $G_{q_m}$ is the question decomposition graph (QDG) associated with $q_m$, which is a direct acyclic graph describing the compositional relationship among questions.
            Moreover, $G_{q_m}$ stores in which manner the questions are decomposited (\ie, the operators in the decomposition program) as the attribute of edges.
            $\hat{a}_m$ denotes the predicted answer for $q_m$.
        }
        \label{fig:method-framework}
    \end{figure*}

    \subsection{Video Question-Answering}

    While the architecture of VidQA methods has undergone significant changes over the years, the essential components of these methods remain the same: video representation, question representation, and video-question aligned representation.
    For the video representation, appearance features~\cite{HeZRS16} and motion features~\cite{XieGDTH17} were commonly used, then the object-level representation was introduced~\cite{JinZG00Z19}.
    For the question representation, most existing works relied on word embeddings~\cite{PenningtonSM14} with RNNs, while BERT~\cite{DevlinCLT19} features became widely used in more recent works~\cite{XiaoYL0JC22, 00040XC22}.
    In the early research, the video-question alignment was implemented using cross-modal attention~\cite{LiSGLH0G19, GaoZSLLMS19} or memory networks~\cite{GaoGCN18, FanZZW0H19}, then graph reasoning~\cite{JiangH20, ParkLS21, Liu0WL21, GuZJHW21, CherianHMR22} became popular.
    Recently, the natural hierarchy in video representation~\cite{LeLV020, GuoZJ0L20, XiaoYL0JC22, PengWG0Z22, abs220705342} received more attention.

    Despite these advancements, existing methods still face challenges in achieving satisfactory levels of compositional consistency~\cite{gandhi2022measuring}.
    To address these challenges, in this paper, we propose a \textit{model-agnostic framework} for compositional reasoning by combining the visual alignment and the answer aggregation to improve current VidQA methods.

    \subsection{Compositional Reasoning in VidQA}

    The practice of decomposing a complex question into simpler questions has been observed in various tasks~\cite{cao2018visual, yang2018hotpotqa}.
    In VidQA, most of earlier efforts~\cite{grunde2021agqa} broke down questions into modular programs that were defined in a neural modular network~\cite{qian2022dynamic} to answer the question.
    AGQA~\cite{grunde2021agqa} explored spatio-temporal scene graphs to represent the programs for VidQA.
    However, such a reasoning program cannot be directly used by existing VidQA methods.
    To address this issue, AGQA-Decomp~\cite{gandhi2022measuring} transferred each reasoning program into several sub-questions and QDG to evaluate the compositional consistency of existing VidQA methods.


    The Neural Modular Network (NMN)-based methods (\eg, DSTN~\cite{qian2022dynamic}) modularized the VidQA task into multiple modules (\emph{e.g.}, \textit{FindObj}, \textit{TemporalFilter}, \emph{etc.}) and generated the reasoning program through a modular policy.
    Although the NMN-based approaches provide perfect interpretability, there still exist three main challenges:
    1) the basic modulars and logic rules have to be pre-defined, making any novel modulars and logic rules incompatible;
    2) compared to conventional neural networks, training NMNs can be more challenging because the learning process optimizes the composition strategy other than the individual modules;
    3) as the number of modules increases, the search space for optimal modules grows exponentially, which hinders its scalability to more complex tasks or larger datasets.

    \subsection{Video Grounding}
    Video grounding~\cite{anne2017localizing, gao2017tall} seeks to identify the most relevant moment in a video based on language queries~\cite{zhang2020regularized, wang2021structured, liu2021context, li2022compositional}, and has received growing attention from downstream video-language tasks~\cite{0004WXJC22, 00040XC22,li2021interventional,anderson2018vision}. 
    Previous works such as IGV~\cite{0004WXJC22} and EIGV~\cite{00040XC22} have focused on differentiating between causal and environment clips in VidQA through a simple grounding indicator and encouraging sensitivity to semantic changes in the causal scene, respectively.
    In contrast to these approaches, our framework hierarchically aligns the question with video clips from object-level, appearance-level, to motion-level to provide a more refined video context along with an answer aggregator.
    This approach enhances both the generalization and compositional reasoning abilities of existing VidQA methods, leading to more effective and accurate models.

    \section{Our Method}
    As described in~\Cref{sec:introduction}, current VidQA models suffer from insufficient compositional reasoning ability.
    Therefore, we propose our VA$^3$ framework, consisting a hierarchical video aligner and a QDG-based answer aggregator.

    \subsection{Model-agnostic VidQA Framework}\label{subsec:model-agnostic-vidqa-framework}
    Our VidQA framework is illustrated in~\Cref{fig:method-framework}.
    For each original question in dataset, it is decomposed (either oracularly or
    with our question decomposition pipeline) into several sub-questions.
    Formally, the original question (regarded as main question) $q_m$ and its decomposed sub-questions $q_{s_i}^m$ form a question cluster on corresponding QDG~$G_{q_m}$.
    First, we introduce a hierarchical video aligner, which selects the question-related video clips by hierarchically interacting the video clips with the question among the object-level, appearance-level and motion-level.
    After that, the questions in the cluster and their corresponding video clips are send into a VidQA model, regarded as $\mathcal{F}:(v, q)\rightarrow \bm{f}_{v, q} \in \mathbb{R}^{h}$, where $h$ is the hidden dimension for joint feature space, to generate the joint feature $\bm{f}_{v,q}$.
    The answer aggregator takes all the joint features from the questions cluster and associates each joint feature with regard to the question node in the QDG.
    By aggregating the information in joint features of each node through QDG, we adjust the question representations and predict the answers.

    \subsection{Video Alignment}
    The structure of video aligner is shown in~\Cref{fig:method-framework}.
    The video aligner takes video-question pair ($v,q$) as input, and gives the video clips that are most relevant to the question.
    Former research on localizing video clips with questions~\cite{0004WXJC22, 00040XC22} failed to use the natural hierarchy of video features, thus limited the representation ability of the aligner.

    In our video aligner, a video $v$ is represented within three-level features: object feature $\bm{F}_o\in \mathbb{R}^{n_c \times n_f \times n_o \times h_v}$, appearance feature $\bm{F}_a\in\mathbb{R}^{n_c\times n_f\times h_v}$ and motion feature $\bm{F}_m \in \mathbb{R}^{n_c\times h_v}$, where $n_c$, $n_f$  and $n_o$ represents the number of clips per video, frames per clip and objects per frame respectively, and $h_v$ is the hidden dimension for each feature vector.
    As these three-level features naturally follow a hierarchical relationship, we designed a hierarchical video aligner to capture them for a better alignment.


    Our hierarchical video aligner follows a bottom-up video-question interaction scheme.
    Starting from $\bm{F}_o$, we aggregate its information among objects with the condition of question, concatenate it with corresponding $\bm{F}_a$, and further aggregate it with other frames in control of question.
    Then, such cross-frame representation is concatenated with $\bm{F}_m$, and interacted with question feature to generate the grounding score.
    During each aggregation, we fuse the video feature and question feature through a transformer layer, where the video feature is regarded as query while question feature serves as the key and value.
    Formally, the aggregation from object feature to appearance feature is
    \begin{align}
        & \bm{F}_o^j\!=\!\text{TF}(\bm{F}_o,\!\bm{F}_q,\!\bm{F}_q); \\
        & \bm{F}_o^a\!=\! \sum_{n_o} \sigma_{n_o}\left(\bm{W}_o\bm{F}_o^j + \bm{b}_o \right) \bm{F}_o^j ;
        \bm{F}_a^c \!=\! [\bm{F}_o^a|| \bm{F}_a];\nonumber
    \end{align}
    where $\sigma_{n_i}$ is the softmax function along $n_i$ dimension, TF is the transformer encoder layer, $\bm{W}_o$ and $\bm{b}_o$ are trainable parameters, and $[\cdot||\cdot]$ denotes the concatenation operator.
    The updated appearance feature $\bm{F}_a^c$ is used to produce aggregated motion feature $\bm{F}_m^c$ in a similar manner.
    Further, we use the $\bm{F}_m^c$ to generate the binary indicator of relevant clip for the video, which can be formulated as
    \begin{align} 
        &\bm{F}_m^j  = \text{TF}(\bm{F}_m^c, \bm{F}_q, \bm{F}_q); \bm{s}_{rel} = \text{MLP}_1(\bm{F}_m^j));   \\ &\bm{s}_{irr} = \text{MLP}_2(\bm{F}_m^j);
        I  = \text{Gumble-Softmax}([\bm{s}_{rel}||\bm{s}_{irr}]),\nonumber
    \end{align}
    where
    \text{MLP} is the multi-layer linear projection.


    Since the ground-truth of aligned video is not applicable in VidQA dataset, we exploit the contrastive learning~\cite{00040XC22} to guide this module.
    Formally, given a video-question pair $(v, q)$ in training data, the indicator specifies the relevant video clips~$\hat{v}_r$ and irrelevant video clips~$\hat{v}_c$ within $v$.
    Thus, the anchor~$\bm{f}_{\hat{v}_r, q}$, positive sample~$\bm{f}_{\hat{v}', q}$, and negative sample~$\bm{f}_{\hat{v}_c, q}$ of the contrastive loss is presented as
    \begin{equation} 
        \bm{f}_{\hat{v}_r,\!q} =\!\mathcal{F}(\hat{v}_r,\!q);  \bm{f}_{\hat{v}',\!q}\!=\!\mathcal{F}(v',\!q); \bm{f}_{\hat{v}_c,\!q}\!=\!\mathcal{F}(\hat{v}_c,\!q),
    \end{equation}
    where $\hat{v}'$ is acquired by replacing $\hat{v}_c$ in $v$ with random sampled clips.
    Therefore, the contrastive loss is defined as
    \begin{equation}
        \mathcal{L}^c_{al} = -\log\frac{\exp(\bm{f}_{\hat{v}_r, q}^T \bm{f}_{\hat{v}', q})}{\exp(\bm{f}_{\hat{v}_r, q}^T \bm{f}_{\hat{v}', q}) + \exp(\bm{f}_{\hat{v}_r, q}^T \bm{f}_{\hat{v}_c, q})}.
    \end{equation}
    Moreover, the answer prediction can be formulated as
    \begin{equation} 
    P(\hat{a}|\hat{v}_r, \!q) \!=\! \sigma(\bm{W}_{o_1} \bm{f}_{\hat{v}_r,q} \!+\! \bm{b}_{o_1}\!),
    \end{equation}
    where $\sigma$ is softmax function and $\bm{W}_{o_1}$ and $\bm{b}_{o_1}$ are the trainable parameters.
    Therefore, the total loss of each video-question pair $(v, q)$ for video aligner can be formulated as
    \begin{equation}
        \mathcal{L}_{al} = \text{CE}(P(\hat{a}|\hat{v}_r, q), a) + \mathcal{L}^c_{al}.
    \end{equation}
    where CE refers to cross entropy and $a$ represents the ground-truth answer for the video-question pair $(v, q)$.



    \subsection{Answer Aggregation}
    Existing VidQA methods predict the answers of different questions independently, which ignores the correlations among questions from the same cluster, leading to insufficient compositional consistency.
    Therefore, we introduce an answer aggregator to supplement the main-question with sub-questions and enhance the compositional consistency.
    Specifically, assume $\{\bm{f}_{q_i, v_i}|i \in \{s_1, \cdots, s_n, m\}\}$ is the video-question joint feature extracted by backbone VidQA model for all questions in QDG $G_{q_m} = (V_{q_m}, E_{q_m})$.
    We explore the graph attention network (GAT) to aggregate the joint feature along the given QDG.
    Formally, given the $k$-th layer of the GAT, the main question $q_m$ and its sub-questions $\{q_{s_1}, \cdots, q_{s_n}\}$, the information aggregation for node associated with $q_i$ is formulates as
    \begin{align} 
            & s(\bm{f}^{k}_{q_i, v_i}, \bm{f}^{k}_{q_j, v_j})  = \bm{a}_k^T \text{LeakyReLU}(\bm{W}_s^k [\bm{f}^{k}_{q_i, v_i}||\bm{f}^{k}_{q_j, v_j}]);\nonumber\\
            &\alpha_{i, j}  = \sigma_j (s(\bm{f}^{k}_{q_i, v_i}, \bm{f}^{k}_{q_j, v_j})); \\
            &\bm{f}^{k+1}_{q_i, v_i}  = \text{ReLU} \left( \sum_{q_j\in \{(q_i, q_j)\in E_{q_m}\}} \alpha_{i,j} \bm{W}_g^k \bm{f}^k_{q_j, v_j} \right),\nonumber
    \end{align}
    where $i, j \in \{s_1, \cdots, s_n, m\}$, and $\bm{a}_k$, $\bm{W}_s^k$ and $\bm{W}_g^k$ are the trainable parameters for the $k$-th layer of GAT.
    Finally, the outputs of all layers is concatenated together and projected to predict the answer, which is formulated as
    \begin{equation} 
        \begin{aligned}
        & \bm{f}_{q_i, v_i}^a = \bm{W}_{o_2} [\bm{f}_{q_i, v_i}^{1}||\cdots||\bm{f}_{q_i, v_i}^{K}] + \bm{b}_{o_2};\\
        & P_{ag}(a_i|v_i, q_i) = \sigma(\bm{f}_{q_i, v_i}^a),
        \end{aligned}
    \end{equation}
    where
    $\bm{W}_{o_2}$ and $\bm{b}_{o_2}$ are trainable parameters.

    Moreover, to enhance the compositional consistency, we introduce an additional contrastive training scheme.
    As the type of relation (\emph{i.e.}, edge) between the questions provides crucial clues in question decompositing and compositional reasoning, we introduce a heuristic prior, where edges with the same type shall have similar representations, and the distance between different types of edges shall be relatively large.
    By leveraging this prior, we can raise the level of abstraction for more accurate and consistent answer reasoning.
    Formally, $\{\bm{f}^e = \bm{W}_e [f_{q_i, v_i}^a || f_{q_j, v_j}^a] + \bm{b}_e|e\in E_{q_m}\}$ denotes the set of node relation representation for edges in graph $G_{q_m}$, where $\bm{W}_e$ and $\bm{b}_e$ are the trainable parameters.
    Moreover, we use $t_e\in T$ to denote the class of edge $e$, where $T$ is the set of edge types.
    For $t \in T$, we use $t^{c} = T / \{t\}$ to represent the complementary set of $t$, and use $e_t$ to denote a random edge sampled from all edges with type $t$.
    Thus, the triplet loss for main question $q_m$ is
    \begin{equation} 
        \mathcal{L}_c^{q_m}\!=\!\mathbb{E}_{e\in E_{q_m}}\!\max\!\left( d(\bm{f}^e,\!\bm{f}^{e_{t_e}})\!-\!d(\bm{f}^e,\!\bm{f}^{e_{t_e^c}})\!+\!m,\!0\right),
    \end{equation}
    where $d(\cdot, \cdot)$ is Eular distance and $m$ is margin.
    Thus, the answer aggregation loss for question cluster $Q = \{q_m, q_{s_1}, \cdots, q_{s_n}\}$ is formulated as
    \begin{equation} 
        \mathcal{L}_{ag}\!=\!\mathcal{L}_c^{q_m}\!+\!\mathbb{E}_{(v_i,q_i)\in \mathcal{D}} \text{CE}(P_{ag}(a_i|v_i,\!q_i),\!a_i),
    \end{equation}
    where $\mathcal{D} = \{(v,q)|q\in Q\}$. 
    Both $\mathcal{L}_{ag}$ and $\mathcal{{L}}_{al}$ are consequently applied during training.

    \subsection{Automatic Question Decomposition Pipeline}\label{subsec:automatic-question-decomposition-pipeline}
    \begin{figure}[t]
        \centering
        \includegraphics[width=0.9\linewidth]{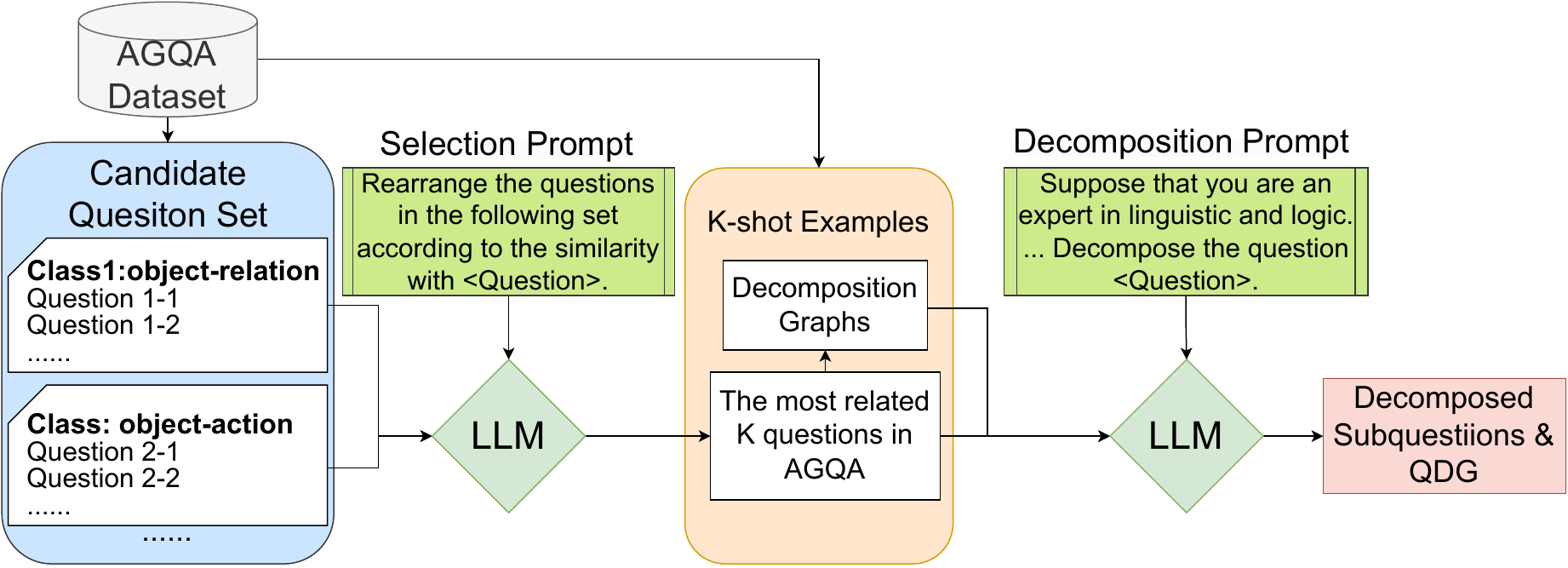}
        \caption{The automatic question decomposition pipeline. The question to be decomposed is denoted as $<$Question$>$.}
        \label{fig:auto-qdp}
    \end{figure}

    For some VidQA dataset (\eg, MSVD and NExT-QA), the QDGs are not applicable as they only provide the main questions.
    To address this issue, we explore an automatic question decomposition pipeline for VidQA data using the knowledge in LLMs.
    Since directly asking the LLM to decompose questions may result in poor results, and random examples could cause unstable quality as the chosen examples may have different compositional structure with the queried question, we proposed the decomposition pipeline as shown in~\Cref{fig:auto-qdp}.
    Firstly, we construct a candidate example set based on the AGQA-Decomp dataset manually, in which each subset consists of a few main questions chosen with a main question type in AGQA-Decomp dataset to cover as many types of main questions as we can.
    Then, we construct a selection prompt which ask the LLM to select the most similar K question, and they form K-shot examples with their QDGs.
    Finally, these K-shot examples are provided to LLM with a decomposition prompt asking the LLM to decompose the target question, resulting in the decomposed sub-questions and corresponding QDGs with better quality.
    {More details and explanations of our pipeline and prompts are in the supplementary material.}

   \section{Metrics}
    \label{sec:metrics}
    Compositional consistency measures whether a method can provide the correct answer for the right reason.
    AGQA-Decomp~\cite{gandhi2022measuring} propose compositional accuracy (CA), right for the wrong reasons (RWR), and Delta (CA-RWA), which offer some insight on it.
    However, as we are to illustrate in~\Cref{subsec:another-view-of-existing-metrics}, they cannot fully reveal the reasoning capabilities, leading to asymmetric and unstable problem.
    To address this issue, we extend them with compositional precision (cP), recall~(cR), and $\text{F}_1$~(c-$\text{F}_1$), along with their negative versions, providing a more comprehensive assessment of reasoning consistency.
    We name $q_j$ as parent question, and $q_i, q_k$ as children question of $q_j$ for QDG edges like $q_i \leftarrow q_j \rightarrow q_k$.
    Note that we only consider the \textbf{1-st} order parent-children relation, and the intermediate sub-questions can be both parent or children regarding the viewpoint.

    \subsection{Another View of Existing Metrics}\label{subsec:another-view-of-existing-metrics}
    CA and RWR are designed to evaluate the accuracy of the parent questions, conditional on whether all their children questions are correct.
    {Let} $\text{N}_-^+$ denote the number of correctly answered main-question with any sub-question incorrect, while $\text{N}_+^-$ denotes the number of falsely answered main question with all sub-questions correct. $\text{N}_-^-$ and $\text{N}_+^+$ is similarly defined.
    Thus, CA and RWR is formulated as
    \begin{equation} 
        \begin{split}
            \text{CA}\!=\!{\text{N}_+^+}/{(\text{N}_+^+\!+\!N_+^-)}\text{; RWR}\!=\!{\text{N}_-^+}/{(\text{N}_-^+\!+\!\text{N}_-^-)},
        \end{split}
    \end{equation}
    and $\text{Delta}\!=\!\text{RWR}\!-\!\text{CA}$.
    Therefore, both CA and $1 - $RWR can be viewed as \textbf{``precisions''}, as the conditions only inspect the correctness of the children questions and missed out the correctness of the parent questions, leading to asymmetric and unstable problems illustrated in~\Cref{tab:mertics-conterexample}.

    For row 1 to row 4, the corresponding model shall have the same compositional consistency, because in all 210 parent questions, 110 of them can be answered for the right reasoning and the opposite.
    However, the CA, RWR and Delta varies significantly among these models, which indicate that the CA-RWR-Delta metrics cannot treat $\text{N}_+^+$, $\text{N}_-^-$, $\text{N}_+^-$ and $\text{N}_-^+$ asymmetrically and cannot correctly identify the compositional consistency.
    Moreover, such failure also lead to instability when facing imbalanced child question accuracy distribution, as shown in row 5 to row 7.
    These models have similar compositional consistency, but the CA, RWR and Delta may vary to the extreme opposite value.


    \begin{table}
        \centering
        \resizebox{\linewidth}{!}{%
        \setlength\tabcolsep{5pt}
        \begin{tabular}{rrrrrrrrrrr}
            \toprule
             &\multicolumn{4}{c}{Data Count}& \multicolumn{4}{c}{Existing Metrics} & \multicolumn{2}{c}{Our Metrics}\\
            \cmidrule(lr){2-5} \cmidrule(lr){6-9} \cmidrule(lr){10-11}
            &$N_+^+$& $N_+^-$& $N_-^+$& $N_-^-$& CA & RWR & Delta & Acc. & c-F$_1$ & Nc-F$_1$\\
            \midrule
            1&100 & 100 & 0 & 10 & 50.00 & 0.00 & -50.00 & 47.61 & 66.67 & 16.67\\
            2&10 & 100 & 0 & 100 & 9.09 & 0.00 & -9.09 & 4.76 & 16.67 & 66.67\\
            3&100 & 0 & 100 & 10 & 100.00 & 90.91 & -9.09 & 95.23 & 66.67 & 16.67\\
            4&10 & 0 & 100 & 100 & 100.00 & 50.00 & -50.00 & 52.38 & 16.67 & 66.67\\
            \midrule
            5&99 & 100 & 1 & 0 & 49.75 & 100.00 & -50.25 & 50.00 & 66.22 & 0.00\\
            6&100 & 99 & 0 & 1 & 50.25 & 0.00 & 50.25 & 50.00 & 66.88 & 0.99\\
            7&99 & 99 & 1 & 1 & 50.00 & 50.00 & 0.00 & 50.00  & 66.44 & 0.98\\
            \bottomrule
        \end{tabular}}
        \caption{The conterexamples. Acc. is parent question accuracy.}
        \label{tab:mertics-conterexample}
    \end{table}

    \subsection{Our Metrics}\label{subsec:our-metrics}

    As described in \Cref{subsec:another-view-of-existing-metrics}, CA and $1 - \text{RWR}$ can both be viewed as \textbf{``precisions''}, thus we denote them as \textbf{consistency precision (cP)} and \textbf{negative consistency precision (NcP)} respectively to simplify the following formulation.
    For a more comprehensive view on compositional consistency, we are to introduce the corresponding ``recalls''.

    \begin{definition}[Consistency Recalls]
        Given a VidQA model $M$, the consistency recall (cR) and negative consistency recall (NcR) is defined as
        \begin{equation} 
            \begin{split}
                \text{cR} = \frac{\text{N}_+^+}{\text{N}_+^+ + \text{N}_-^+ }, \quad \text{NcR} = \frac{\text{N}_-^-}{\text{N}_-^- + \text{N}_+^-}.
            \end{split}
        \end{equation}
    \end{definition}


    These metrics condition on the correctness of main questions.
    Clearly, neither cP and cR can represent the compositional consistency solely, since they only take part of the condition into consideration.
    To combine both perspectives, we introduce their weighted harmonic mean, \ie, consistency F-scores for a robust and symmetric representation.

    \begin{definition}[Consistency F-Score]\label{def:comp-F-score}
        Given a VidQA model $M$, the consistency F-score (c-$\text{F}_\beta$) and negative consistency F-score (Nc-$\text{F}_\beta$) is defined as
        \begin{equation} 
            \begin{split}
            \text{c-$\text{F}_\beta$}\!=\!\frac{(1\!+\!\beta^2)\text{cP}\!\cdot\!\text{cR}}{\beta^2\text{cP}\!+\!\text{cR}}\text{; Nc-$\text{F}_\beta$}\!=\!\frac{(1\!+\!\beta^2)\text{NcP}\!\cdot\!\text{NcR}}{\beta^2\text{NcP}\!+\!\text{NcR}}.
            \end{split}
        \end{equation}
    \end{definition}

    In such definition, $\beta$ is a hyper-parameter to balance cP and cR.
    To measure the two kinds of error in a equal weight, we set $\beta = 1$, and thus use c-$\text{F}_1$ to measure the compositional consistency of models.
    Such metric considers the compositional consistency in a symmetric manner, raising a comprehensive evaluation on model's ability.
    As shown in~\Cref{tab:mertics-conterexample}, our c-$\text{F}_1$ and Nc-$\text{F}_1$ metrics raises more reasonable evaluations (row 1 to row 4), and is also more stable facing extreme cases (row 5 to row 7).
    Note that although c-$\text{F}_1$ and Nc-$\text{F}_1$ provides a balanced view of compositional consistency, we still need to the accuracy, cP, and cR for a detailed analysis regarding prediction ability and compositional bias.
    \textbf{More analysis and comparison between our metrics and original ones are in the supplementary.}

    \section{Experiments}
    \begin{table*}[ht]
    \centering
    \resizebox{\textwidth}{!}{%
    \setlength\tabcolsep{2.5pt}
    \begin{tabular}{lllllllllllll}
    \toprule
     & \multicolumn{3}{c}{Main Accuracy} & \multicolumn{3}{c}{Sub Accuracy} & \multicolumn{6}{c}{Compositational Consistency} \\     \cmidrule(lr){2-4} \cmidrule(lr){5-7} \cmidrule(lr){8-13}
              & Open  & Binary & \textbf{All}   & Open  & Binary & \textbf{All}   & cP  & cR  & \textbf{c-F${\bm{_1}}$}  & NcP  & NcR  & \textbf{Nc-F${\bm{_1}}$}  \\ \midrule
    HME          & 36.29 & 51.41 & \textbf{41.59} & 29.68 & 71.73 & \textbf{56.59} & 53.06 & 46.02 & \textbf{49.29} & 56.57 & 63.33 & \textbf{59.76}\\
    HGA          & 41.18 & 56.62 & \textbf{46.61} & 36.03 & 73.11 & \textbf{59.75} & 64.02 & 57.85 & \textbf{60.78} & 57.35 & 63.55 & \textbf{60.29}\\
    HQGA         & 41.05 & 50.40 & \textbf{44.34} & 33.22 & 70.03 & \textbf{56.77} & 54.49 & 38.55 & \textbf{45.16} & 53.89 & 69.04 & \textbf{60.53}\\
    VA$^3$(HME)  & 39.91\mygreen{$^{+3.62}$} & 52.26\mygreen{$^{+0.85}$} & \textbf{44.30}\mygreen{$^{+2.71}$} & 35.23\mygreen{$^{+5.55}$} & 73.56\mygreen{$^{+1.83}$} & \textbf{59.75}\mygreen{$^{+3.16}$} & 57.78\mygreen{$^{+4.72}$} & 48.66\mygreen{$^{+2.64}$} & \textbf{52.83}\mygreen{$^{+3.54}$} & 55.81\myred{$^{-0.76}$} & 64.58\mygreen{$^{+1.25}$} & \textbf{59.87}\mygreen{$^{+0.11}$}\\
    VA$^3$(HGA)  & 43.04\mygreen{$^{+1.86}$} & 56.82\mygreen{$^{+0.20}$} & \textbf{47.88}\mygreen{$^{+1.27}$} & 42.18\mygreen{$^{+6.15}$} & 74.69\mygreen{$^{+1.58}$} & \textbf{62.98}\mygreen{$^{+3.23}$} & 67.52\mygreen{$^{+3.50}$} & 60.38\mygreen{$^{+2.53}$} & \textbf{63.75}\mygreen{$^{+2.97}$} & 57.47\mygreen{$^{+0.12}$} & 64.83\mygreen{$^{+1.28}$} & \textbf{60.93}\mygreen{$^{+0.64}$}\\
    VA$^3$(HQGA) & 42.35\mygreen{$^{+1.30}$} & 51.53\mygreen{$^{+1.13}$} & \textbf{45.57}\mygreen{$^{+1.23}$} & 35.28\mygreen{$^{+2.06}$} & 74.01\mygreen{$^{+3.98}$} & \textbf{60.06}\mygreen{$^{+3.29}$} & 56.02\mygreen{$^{+1.53}$} & 42.50\mygreen{$^{+3.95}$} & \textbf{48.33}\mygreen{$^{+3.17}$} & 55.26\mygreen{$^{+1.37}$} & 68.04\myred{$^{-1.00}$} & \textbf{60.99}\mygreen{$^{+0.46}$}\\
    \bottomrule
    \end{tabular}%
    }
    \caption{The comparison with baseline methods on AGQA-Decomp~\cite{gandhi2022measuring}. The overall measurements are highlighted in bold, and the improvements of our framework are highlighted as superscript.}
    \label{tab:main_balanced}
    \end{table*}

    \subsection{Experiment Setting}
    \textbf{Dataset}
    As described in~\Cref{sec:introduction}, we conduct our experiment on AGQA-Decomp benchmark, which extends AGQA~2.0 by decomposing each question into several sub-questions with a QDG, and evaluates the compositional reasoning ability by supplying extensive challenging complex questions with their decomposed sub-questions and answers.
    Moreover, we test the improvement on MSVD and NExT-QA dataset to verify the applicability of our VA$^3$ framework and question decomposition pipeline.

    \noindent\textbf{Baselines and Metrics}
    We conduct experiment on all three categories of VidQA mothods, \ie, memory-based, graph-based and hierarchy-based methods.
    Specifically, we choose HME~\cite{FanZZW0H19}, HGA~\cite{JiangH20} and HQGA~\cite{XiaoYL0JC22} as the baseline methods from each category respectively.
    For the evaluation metrics, we measure the open-ended, binary, and overall accuracy for main questions and sub-questions.
    Moreover, to illustrate the improvement in terms of compositional consistency, we evaluate the cR, cP, c-F$_1$, NcR, NcP and Nc-F$_1$.
    More detailed settings are in supplementary material.

    \subsection{Main Results}
    The result of our framework with various baselines is shown in~\Cref{tab:main_balanced}.
    Compared the 1-st to 3-rd row with the 4-th to 6-th row correspondingly, our framework outperforms all baseline models, including memory-, graph- and hierarchy-based models, significantly in terms of both accuracy and compositional consistency.
    For accuracy, the overall main question accuracy improves 1.23\% to 2.71\%, while the sub-question accuracy raises 3.16\% to 3.29\%.
    The accuracy improvement on sub-questions are usually more than that on main questions.
    For video aligner, it is more hard to align the corresponding video clips for main questions, besides, for answer aggregator, it is also more challenge to aggregate all the sub-questions to deduce the main question, leading to such improvement gap.
    Moreover, the accuracy improvements on binary and open-ended questions are not equal.
    The reasons include the the following two aspects:
    1) the video aligner helps open-ended questions more since they are more sensitive to irrelevant clips as they have to choose answer from a much larger candidate set than binary questions, and may be mislead by the actions in irrelevant clips more easily;
    2) the open-ended questions provide and receive more severe information in answer aggregation, making the answer aggregator contribute more on them.
    

    Moreover, the compositional consistency also raises significantly compared to the baseline models.
    Specifically, the c-F$_1$ significantly improves 2.97\% to 3.54\%, while the Nc-F$_1$ raises 0.11\% to 0.64\%.
    The c-F$_1$ indicates how much the model answers \textbf{correctly} with \textbf{correct} inference, therefore, the c-F$_1$ is the most important overall measurement for the reasoning ability of models.
    The significant improvement in terms of c-F$_1$ further indicates that our framework does help the VidQA models reasoning correctly and consistently.
    And the Nc-F$_1$, which measures if the model is still consistent even when the answer is incorrect, shows that our framework also slightly helps improve the overall consistency on the incorrect main questions.
    Furthermore,  we can also find that the the improvement on c-F$_1$ are naturally more significant that on Nc-F$_1$ as our constraint on answer aggregation mainly focus on correctly deducing the main question based on the correct sub-questions.

    \begin{table}[t]
    \centering
    \resizebox{\linewidth}{!}{%
    \setlength\tabcolsep{2.5pt}
    \begin{tabular}{lllllll}
    \toprule
     & \multicolumn{3}{c}{Novel Comp. Setting} & \multicolumn{3}{c}{More Comp. Step Setting} \\
    \cmidrule(lr){2-4} \cmidrule(lr){5-7}
              & Accuracy  & c-F$_1$ & Nc-F$_1$ & Accuracy  & c-F$_1$ & Nc-F$_1$  \\ \midrule
    HME          & {31.54} & {35.88} & {68.94} & {44.28} & {48.32} & {63.71}\\
    HGA          & {33.40} & {35.44} & {65.18} & {47.26} & {48.42} & {63.45}\\
    HQGA         & {34.21} & {38.91} & {67.96} & {46.64} & {50.65} & {62.97}\\
    VA$^3$(HME)  & {33.45}\mygreen{$^{+1.91}$} & {38.26}\mygreen{$^{+2.38}$} & {69.08}\mygreen{$^{+0.14}$} & {46.07}\mygreen{$^{+1.79}$} & {49.87}\mygreen{$^{+1.55}$} & {64.01}\mygreen{$^{+0.30}$}\\
    VA$^3$(HGA)  & {35.27}\mygreen{$^{+1.87}$} & {40.47}\mygreen{$^{+5.03}$} & {65.33}\mygreen{$^{+0.15}$} & {48.38}\mygreen{$^{+1.12}$} & {51.08}\mygreen{$^{+2.66}$} & {63.58}\mygreen{$^{+0.13}$}\\
    VA$^3$(HQGA) & {36.33}\mygreen{$^{+2.12}$} & {40.76}\mygreen{$^{+1.85}$} & {68.20}\mygreen{$^{+0.24}$} & {47.91}\mygreen{$^{+1.27}$} & {51.75}\mygreen{$^{+1.10}$} & {63.26}\mygreen{$^{+0.29}$}\\
    \bottomrule
    \end{tabular}%
    }
    \caption{The comparison with baseline methods on the AGQA-Decomp~\cite{gandhi2022measuring} \textit{novel composition} setting and \textit{more composition step} setting~\cite{grunde2021agqa}. Comp. is the abbreviation for compositional. The improvements of our framework are highlighted as superscript.}
    \label{tab:main_more_step}
    \end{table}

    \subsection{Generalization Ability}

    We further test the improvement of our framework when generalizing to new situations on two extra settings, \ie \textit{novel composition} and \textit{more composition step}~\cite{grunde2021agqa}.
    The \textit{Novel composition} setting tests if models can generalize to unseen composition types, and the \textit{more composition step} setting tests the generalization ability when facing more complex questions than training.
    The results are in~\Cref{tab:main_more_step}.

    When facing novel composition setting, the clear accuracy drop compared to~\Cref{tab:main_balanced} indicates generalizing to novel composition setting is much harder for VidQA models than the standard setting.
    However, our framework is still capable of significantly boosting baseline methods under this challenging setting.
    The main question accuracy raises 1.87\% to 2.12\%, while the c-F$_1$ improves 1.85\% to 5.03\% and the Nc-F$_1$ improves 0.14\% to 0.24\%, on different baselines, implying that our framework provides better generalization ability on unseen composition types, in terms of both accuracy and compositional consistency.

    For the more composition step setting, our framework still significantly improves the baseline methods on both the accuracy and compositional consistency, indicating our framework is effective when generalizing to more complex questions.
    In detail, there is a 1.12\% to 1.76\% accuracy improvement for main questions, while the c-F$_1$ improves 1.10\% to 2.66\%  while the Nc-F$_1$ raises 0.13\% to 0.30\%.
    
    \begin{table}[t]
    \centering
    \resizebox{\columnwidth}{!}{%
    \setlength\tabcolsep{2.5pt}
    \begin{tabular}{clcccccccc}
    \toprule
     &  & \multicolumn{3}{c}{Main Accuracy} & \multicolumn{3}{c}{Sub Accuracy} & \multicolumn{2}{c}{Consistency} \\
    \cmidrule(lr){3-5}\cmidrule(lr){6-8}\cmidrule{9-10}
     &  & Open & Binary & All & Open & Binary & All & c-F$_1$ & Nc-F$_1$ \\ \midrule
    \multicolumn{2}{c}{HME} & 36.29 & 51.41 & 41.59 & 29.68 & 71.73 & 56.59 & 49.29 & 59.76 \\ \midrule
    \multirow{2}{*}{VA.}
    & EIGV~\cite{00040XC22} & 38.84 & 51.48 & 43.29 & 31.77 & 72.42 & 57.77 & 48.05 & 59.31 \\
    & Our Aligner & 39.49 & 51.50 & 43.72 & 32.27 & 72.81 & 58.20 & 49.21 & 59.71 \\
    \midrule
    \multirow{2}{*}{AA.} & +AA. & 38.21 & 51.35 & 42.83 & 33.78 & 72.33 & 58.44 & 52.19 & 59.73 \\
     & +AA. + $\mathcal{L}_c^{q_m}$ & 38.81 & 52.05 & 43.46 & 34.12 & 72.91 & 58.93 & 53.49 & 60.05 \\
     \midrule
    \multicolumn{2}{c}{VA$^{3}$(HME)} & 39.91 & 52.26 & 44.30 & 35.23 & 73.56 & 59.75 & 52.83 & 59.87 \\ \bottomrule
    \end{tabular}%
    }
    \caption{The ablation study on our Video Aligner and Answer Aggregator. VA. is video aligner and AA. is answer aggregator.}
    \label{tab:ablation}
    \end{table}
 
    \begin{figure*}[ht]
        \centering
        \includegraphics[width=0.95\linewidth]{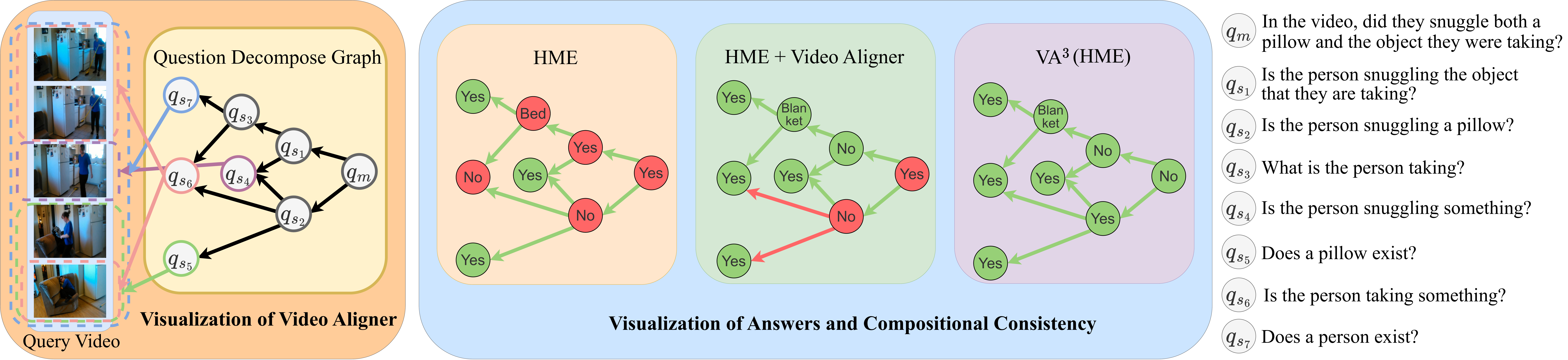}
        \caption{Quantitive results of Video Aligner and the visualization of improvements on accuracy and compositional consistency brought by our modules. Best viewed in color and zoom in. \textbf{More visualizations and explanations are in the supplementary material.}}
        \label{fig:expr-visulization}
    \end{figure*}
    
    \subsection{Ablation Study}\label{subsec:ablation}
    To measure the effectiveness and necessity of our modules, we conduct ablation studies in this section.
    To clearly reveal the contribution of each module and loss, we choose HME~\cite{FanZZW0H19} as the baseline method in this section, since the improvement over HME is the largest among all three baselines.
    For the video aligner, we test its improvement over the EIGV~\cite{00040XC22} aligner brought by the hierarchy structure.
    For the answer aggregator, we test how much it boosts the original model even without our contrastive loss $\mathcal{L}_c^{q_m}$, then measure the improvement of applying $\mathcal{L}_c^{q_m}$ on answer aggregator.
    The results are summarized in~\Cref{tab:ablation}.

    By comparing row 3 with rows 1 and 2, we can conclude that the video aligner substantially improves the accuracy of both main questions and sub-questions, but helps little on compositional consistency.
    This is reasonable since video aligners do not exploit the relations between main questions and sub-questions, thus cannot improve the compositional consistency among them.
    Moreover, the comparison between row 3 and row 2 shows that our aligner outperforms the video grounding module in EIGV due to its hierarchical structure, which could effectively extract information from different level of video feature and align them with the question.
    As for the answer aggregation module, by comparing row 4 with row 1, we can infer that answer aggregator, even without the contrastive loss $\mathcal{L}_c^{q_m}$, can significantly improve the accuracy and compositional consistency, as the information exchange along main-sub questions relation helps both correct answering and ensuring their consistency.
    Further, as the comparison between row 4 and row 5 implies, the contrastive loss $\mathcal{L}_c^{q_m}$ further improves the accuracy and compositional consistency.
    Finally, the improvement of row 6 over row 3 and 5 indicates that our combination of video aligner and answer aggregator is mutually beneficial on both accuracy and compositional consistency, proving the effectiveness of our framework.

    \subsection{Quailititave Study}
    In~\Cref{fig:expr-visulization}, we firstly visualize the result of video aligner.
    As the left part of~\Cref{fig:expr-visulization} shows, our video aligner successfully aligns the related video clips (denoted by the dotted boxes) with the corresponding questions.
    Therefore, the VidQA backbones can use more accurate information to improve the accuracy of both main questions and sub-questions.
    Moreover, we also visualize the predicted answers of HME, HME with video aligner, and VA$^3$(HME) respectively.
    The original model failed to answer the main question correctly due to several factual errors in sub-questions and the potential reasoning failure.
    With the help of video aligner, we may reduce the irrelevant noise by only processing the most correlated clips, thus eliminating several factual errors (\ie, $q_{s_1}, q_{s_3}$ and $q_{s_6}$), but may not help on the main question since the whole video is fatal in generating its answer.
    Moreover, the video aligner may not correct the compositional reasoning failure as it does not use inter-question relation information.
    However, with the help of the answer aggregator, we can correct the reasoning failure by aggregating the information from sub-questions (\eg, correct the answer of $q_{s_2}$  by aggregating video-question joint features of $q_{s_4}, q_{s_5}$ and $q_{s_6}$), and deduce the correct answer of main questions with higher compositional consistency.

    \subsection{Applicability}\label{subsec:applicability}

    \begin{table}[t]
    \centering
    \resizebox{\linewidth}{!}{%
    \setlength\tabcolsep{2pt}
    \begin{tabular}{lllllll}
    \toprule
     & HME  & HGA & HQGA & VA$^3$(HME)  & VA$^3$(HGA) & VA$^3$(HQGA)  \\
    \midrule
    MSVD          & 33.75 & 36.71 & 41.23 & \textbf{38.51}\mygreen{$^{+4.76}$} & \textbf{41.24}\mygreen{$^{+4.53}$} & \textbf{44.46}\mygreen{$^{+3.23}$}\\
    NExT-QA      & 48.72 & 50.04 & 51.65 & \textbf{53.23}\mygreen{$^{+4.51}$} & \textbf{54.11}\mygreen{$^{+4.07}$} & \textbf{55.23}\mygreen{$^{+3.58}$}\\
    \bottomrule
    \end{tabular}%
    }
    \caption{The comparison with baseline methods on MSVD and NExT-QA datasets. Improvements are highlighted as superscripts.}
    \label{tab:applicability}
    \end{table}

    To verify that our framework is generally applicable, we further apply our framework on MSVD~\cite{XuZX0Z0Z17} and NExT-QA~\cite{XiaoSYC21} datasets extended by our automatic question decomposition pipeline.
    The results are summarized in~\Cref{tab:applicability}.
    By comparing the 1st to 3rd columns with 4th to 6th columns in the table, we could find that our framework, with our automatic decomposition pipeline, significantly boosts the performance of baselines on both MSVD and NExT-QA dataset.
    Specifically, the overall accuracy improves 3.23\% to 4.76\% on MSVD, while improves 3.58\% to 4.51\% on NExT-QA, which indicates that with the help of our question decomposition pipeline, our framework can still raise the performance significantly even on datasets which originally do not have QDGs, further implying the applicability of our framework in real world scenarios.
    
    \section{Conclusion}
    In this work, we have focused on the VidQA from interpretability and proposed a model-agnostic align-and-aggregate framework for VidQA.
    It firstly aligns the video representation towards both main question and sub-questions, then aggregates the video-question joint representation through the QDG.
    Further, we have revisited the compositional consistency metrics and have proposed more comprehensive c-F scores.
    Extensive experiments on various VidQA models have revealed that our framework improves both compositional consistency and accuracy significantly, leading to more interpretable VidQA models.

\section*{Acknowledgments}
The work was supported by the National Natural Science Foundation of China (Grant No. 62076162), the Shanghai Municipal Science and Technology Major/Key Project, China (Grant No. 2021SHZDZX0102).

{
\small
\bibliographystyle{ieeenat_fullname}
\bibliography{main}
}


\end{document}